\documentclass[runningheads]{llncs}
\usepackage[T1]{fontenc}
\usepackage{graphicx, amsmath, amssymb, amsfonts, caption, subcaption, multirow, overpic, textpos}
\usepackage{booktabs}
\usepackage{placeins}
\usepackage{bm}
\usepackage[misc]{ifsym} 
\usepackage[colorlinks=true,allcolors=blue]{hyperref}
\begin{document}
\title{DGMIL: Distribution Guided Multiple Instance Learning for Whole Slide Image Classification}
\titlerunning{Distribution Guided Multiple Instance Learning}

\author{Linhao Qu$^*$, Xiaoyuan Luo$^*$, Shaolei Liu, Manning Wang$^{(\textrm{\Letter})}$, Zhijian Song$^{(\textrm{\Letter})}$}
\authorrunning{Qu et al.}

\institute{Digital Medical Research Center, School of Basic Medical Science, 
Shanghai Key Lab of Medical Image Computing and Computer Assisted Intervention, Fudan University, Shanghai 200032, China.
 \{lhqu20, xyluo19, slliu, mnwang, zjsong\}@fudan.edu.cn}

\renewcommand{\thefootnote}{}
\footnotetext{$^*$Linhao Qu and Xiaoyuan Luo are contributed equally to this work.}
\footnotetext{$^*$Code is available at https://github.com/miccaiif/DGMIL.}

\maketitle              
\begin{abstract}
Multiple Instance Learning (MIL) is widely used in analyzing histopathological Whole Slide Images (WSIs). However, existing MIL methods do not explicitly model the data distribution, and instead they only learn a bag-level or instance-level decision boundary discriminatively by training a classifier. In this paper, we propose DGMIL: a feature distribution guided deep MIL framework for WSI classification and positive patch localization. Instead of designing complex discriminative network architectures, we reveal that the inherent feature distribution of histopathological image data can serve as a very effective guide for instance classification. We propose a cluster-conditioned feature distribution modeling method and a pseudo label-based iterative feature space refinement strategy so that in the final feature space the positive and negative instances can be easily separated. Experiments on the CAMELYON16 dataset and the TCGA Lung Cancer dataset show that our method achieves new SOTA for both global classification and positive patch localization tasks.
\keywords{Histopathological images  \and Multiple Instance Learning.}
\end{abstract}
\section{Introduction}
Histopathological image analysis is important for clinical cancer diagnosis, prognostic analysis and treatment response prediction \cite{20}. In recent years, the emergence of digital Whole Slide Images (WSIs) facilitates the application of deep learning techniques in automated histopathological image analysis \cite{3,4,15,16}. Two main challenges exist in deep learning-based WSI analysis. First, WSIs have extremely high resolutions (typically reaching 50,000 $\times$ 50,000 pixels) and thus are unable to be directly fed into deep learning models. Therefore, WSIs are commonly divided into small patches for processing. Secondly, fine grained manual annotations are very expensive and time consuming because of its big size, so patch-wise labels are usually not available, and only the labels of each WSI are known, leading to the inaccessibility of traditional supervised learning methods \cite{2,12}. For these reasons, Multiple Instance Learning (MIL), an effective weakly supervised paradigm has become mainstream technique for deep learning-based WSI analysis, where each WSI is considered as a bag containing many patches (also called instances) \cite{1,5,12,19,22,23,24,25}. If a WSI (bag) is labeled positive, then at least one patch (instance) in it is positive. On the contrary, if a WSI is negative, all patches in it are negative. Typically, WSI image analysis has two main objectives: global classification and positive patch localization. The first objective is to accurately classify each WSI into positive or negative and the second objective is to accurately classify each patch in a positive slide into positive or negative. In line with many recent studies \cite{3,6,7,8,9}, our approach is dedicated to accomplish both tasks.

Existing MIL methods for WSI analysis can be broadly classified into embedding-based methods \cite{6,8,13,9,21} and key-instance-based \cite{3,7} methods. In embedding-based methods, patches in a WSI are first mapped to fixed-length feature embeddings, which are aggregated to obtain the bag-level feature embedding. Then, a bag-level classifier is trained in a supervised way for slide-level classification. Aggregation parameters, such as attention coefficients are used to measure the contribution of different patches to the bag classification and thus to accomplish the patch-level classification task. The embedding-based methods focus on extracting and aggregating features of each instance. The approaches for feature aggregation in existing studies include attention mechanism \cite{6,8}, graph convolution network \cite{13}, masked non-Local operation \cite{9} and Transformer \cite{12,17,18}. Among them, most methods train a feature extractor and the feature aggregator end-to-end, while pretrained models are utilized for feature extraction in \cite{9} and \cite{13}.
In the key-instance-based methods \cite{3,7}, some key patches are first selected and assigned pseudo-labels to train a patch-level classifier in a supervised manner, and both the key-patches with their pseudo-labels and the classifier are updated iteratively. During inference, the trained patch-level classifier can be directly used for patch-level classification and the classification results of all patches in a slide can be aggregated to perform slide-level classification.

We observed that existing methods do not explicitly model the data distribution, and instead they only learn a bag-level or instance-level decision boundary discriminatively by training a classifier. However, since the weak slide-level labels can only provide limited supervisory information, their performance is still limited. For example, in embedding-based methods, the model is mainly trained discriminatively by the bag-level loss so the model is not motivated to make accurate predictions for all instances after identifying some significant instances to accomplish bag classification. In key-instance-based methods, the positive pseudo labels may be wrong or the selected instances may be insufficient, both of which will result in an inferior classifier.

On the basis of the above observation, we propose DGMIL: a feature distribution guided deep MIL framework for WSI classification and positive patch localization. Instead of designing complex discriminative networks, we reveal that the inherent feature distribution of histopathological image data can serve as a very effective guide for classification. Therefore, different from existing MIL methods, we do not concentrate on designing sophisticated classifiers or aggregation methods, but focus on modeling and refining the feature space so that the positive and negative instances can be easily separated. To this end, we propose a cluster-conditioned feature distribution modeling method and a pseudo label-based iterative feature space refinement strategy. Specifically, we use the self-supervised masked autoencoders (MAE) \cite{10} to map all instances into an initial feature space and iteratively refine it for better data distribution. In each iteration, we first cluster all the instances from negative slides in the training set and calculate positive scores for all instances from both negative and positive slides in the training set according to their Mahalanobis distance to each cluster. Then, the instances with the lowest and the highest positive scores are assigned negative and positive pseudo-labels, respectively, and are used to train a simple linear projection head for dynamic refinement of the feature space. This process iterates until convergence. For testing, we map all test instances to the final refined feature space and calculate the positive scores of each instance for the patch-level classification task. For bag-level classification, we only use the simple mean-pooling approach to aggregate the positive scores of all instances in a bag.

The main contributions of this paper are as follows:

\noindent$\bullet $ We propose DGMIL: a feature distribution guided deep MIL framework for WSI classification and positive patch localization. Instead of designing complex discriminative network architectures, we reveal that the inherent feature distribution of histopathological image data can serve as a very effective guide for instance classification. To the best of our knowledge, this is the first work that explicitly solves a deep MIL problem from a distribution modeling perspective.

\noindent$\bullet $ We propose a cluster-conditioned feature distribution modeling method and a pseudo label-based iterative feature space refinement strategy so that in the final feature space the positive and negative instances can be easily separated.

\noindent$\bullet $ Experiments on the CAMELYON16 dataset and the TCGA Lung Cancer show that our method achieves new SOTA for both global classification and positive patch localization tasks. Codes will be publicly available.

\section{Method}
\subsection{Problem Formulation}
Given a dataset consisting of $N$ WSIs $W=\left\{W_1,W_2,\ldots,W_N\right\}$, and each slide $W_i$ has the corresponding label $Y_i\in\left\{0,1\right\},\ i=\left\{1,2,...N\right\}$. Each slide $W_i$ is divided into small patches $\left\{p_{i, j}, j=1,2, \ldots \mathrm{n}_{i}\right\}$ without overlapping, and $n_i$ is the number of patches in the $ith$ slide.
In the MIL formulation, all patches $\left\{p_{i, j}, j=1,2, \ldots \mathrm{n}_{i}\right\}$ from a slide $W_i$ constitute a bag, and the bag-level label is the label $Y_i$ of $W_i$. Each patch is called an instance, and the instance label $y_{i,j}$ has the following relationship with its corresponding bag label $Y_i$:
\begin{equation}
    Y_{i}=\left\{\begin{array}{cc}
        0, & \text { if } \sum_{j} y_{i, j}=0 \\
        1, & \text { else }
        \end{array}\right.  \label{eq1}
\end{equation}
\begin{figure*}[htbp]
    \centering
    \includegraphics[scale=0.20]{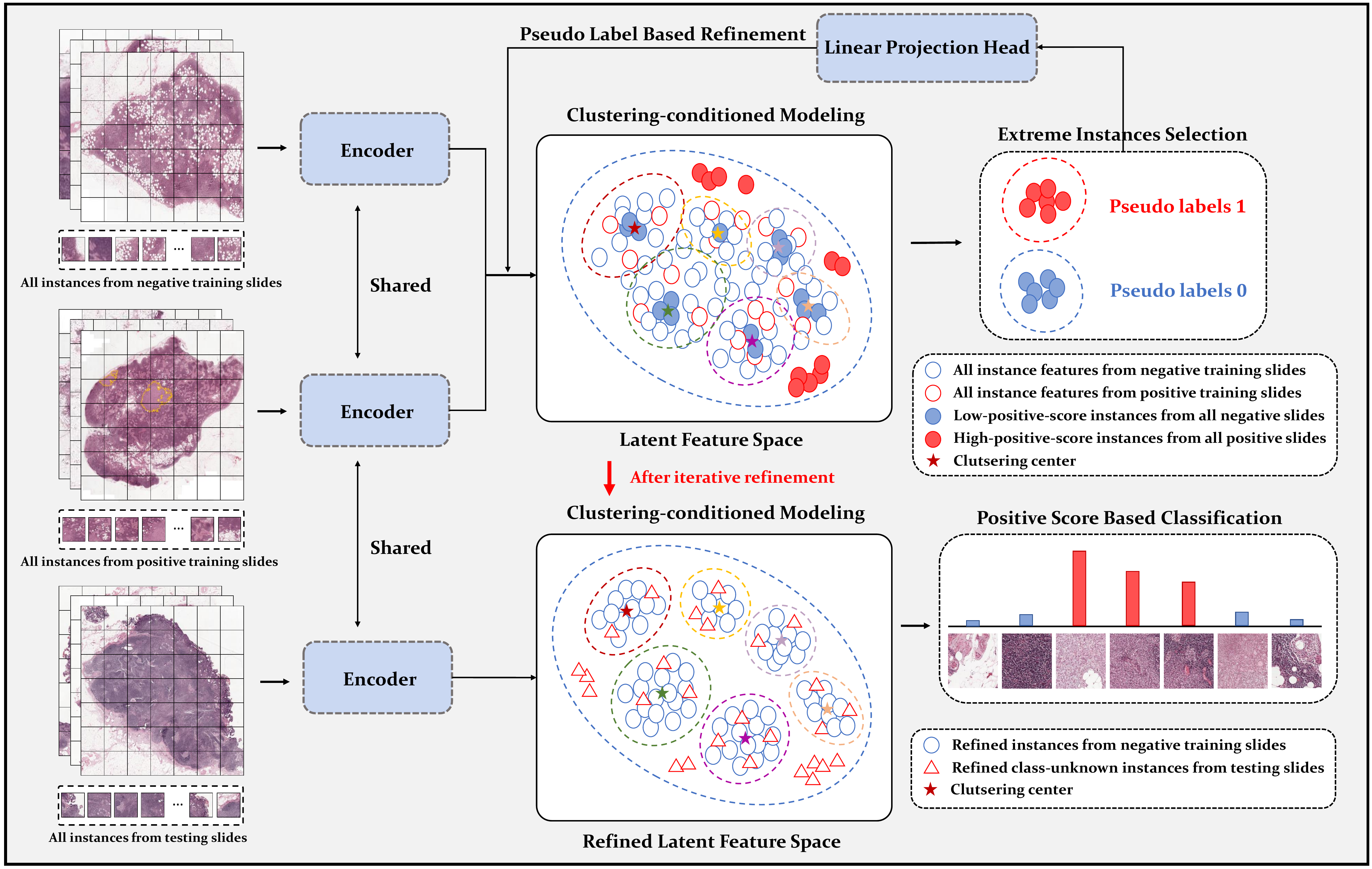}
    \caption{The overall framework of DGMIL. (Best view in color.)}
    \label{figure1}
\end{figure*}

That is, the labels of all instances in a negative bag are negative, while there is at least one positive instance in a positive bag but which ones are negative is unknown. We define a feature extractor $f:p_{i,j}\rightarrow z_{i,j}$, where $z_{i,j}\in R^d$, which maps the instances $p_{i,j}$ to a d-dimensional latent feature vector $z_{i,j}$. This feature extractor is typically parameterized by a deep neural network. There are significant differences in cell morphology between tumor and normal tissues, and therefore their distribution in the feature space should also be significantly different if a proper latent space can be found. We assume that the feature vectors of all negative instances and all positive instances are derived from the distribution $P_{z_{i,j}}^{neg}$ and $P_{z_{i,j}}^{pos}$, respectively, and the two distributions are significantly different from each other. On this basis, our goal is to model the feature space properly to make the negative and positive instances easily separable in the feature space. Since our methods are based on the features of instances, the instances we mention later in this paper all refer to their feature vectors.

\subsection{Framework Overview}
Fig. \ref{figure1} illustrates the overall framework of our proposed DGMIL. Specifically, we use masked autoencoders to perform self-supervised learning to train the Encoder to map all instances to an initial latent feature space, which will be iteratively refined. The iterative feature space refinement process is described in detail in section 2.5, and the final refined latent feature space is used for instance classification. During inference, we map a test instance from a test slide to the refined latent feature space and calculate its positive score for instance-level classification. For bag-level classification, we only use the simple mean-pooling approach to aggregate the positive scores of all instances in a bag. Since both the training and the inference are based on each independent instance (i.e. without using its position information in the slide), our method is of permutation invariance. 

\subsection{Self-supervised Masked Autoencoders for Feature Space Initialization}
It is crucial to learn an instance-level feature representation that is suitable for distribution modeling. We propose to use the state-of-the-art self-supervised learning framework masked autoencoders (MAE) \cite{10} to extract the feature representations and initialize the feature space. MAE learns a robust feature representation by randomly masking a high proportion of patches of the input image and reconstructing the missing pixels using Transformer. We first train the MAE with all instances from the negative and positive slides, and then use the trained encoder as an instance-level feature extractor and map all instances in the training bags into the initial feature space.

\subsection{Cluster-conditioned Feature Distribution Modeling}
We propose a feature distribution modeling method based on K-means clustering and Mahalanobis distance. Specifically, we first use the K-means algorithm to cluster all instances from negative slides in the training set into $M$ clusters, where each cluster is denoted as $C_m$. Next, we compute the positive scores $s_{i,j}$ using the Mahalanobis distance for all instances from both negative slides and positive slides in the training set,
\begin{equation}
    s_{i, j}=\min _{m} D\left(z_{i, j}, C_{m}\right)=\min _{m}\left(z_{i, j}-\mu_{m}\right)^{T} \sum_{m}^{-1}\left(z_{i, j}-\mu_{m}\right)  \label{eq2}
\end{equation}
where $D(\cdot)$ denotes the Mahalanobis distance, and $\mu_m$ and $\Sigma_m$ are the mean and covariance of all instances in the cluster. It can be seen that the positive score of an instance is actually the minimum of the distances from the instance to each cluster. A higher positive score indicates a higher probability that the instance is positive, and vice versa.

\subsection{Pseudo Label-Based Feature Space Refinement}
The direct use of the initial feature space based on MAE does not model the distribution of positive instances and negative instances well because the training of MAE is completely self-supervised and the bag-level supervision is not utilized. Therefore, we further propose a pseudo label-based feature space refinement strategy to refine it. 

This feature space refine strategy is an iterative process. In each iteration, we first cluster all the instances from the negative slides in the training set using the K-means algorithm and calculate the positive scores of all instances from both positive and negative slides. A proportion of the instances with the highest positive scores from the positive slides and a proportion of instances with the lowest positive scores from the negative slides are called extreme instances, and they are assigned pseudo labels 1 and 0, respectively. With these extreme instances and their pseudo labels, we train a simple binary classifier consisting of a one-FC-layer Linear Projection Head and a one-FC-layer Classification Head in a supervised manner. Finally, we utilize the Linear Projection Head to remap the current instance feature into a new feature space to achieve the feature space refinement. The above process of feature space refinement iterates until convergence.

\section{Experimental Results}

\subsection{Datasets}
\textbf{CAMELYON16 dataset.} CAMELYON16 dataset \cite{11} is a widely used publicly available dataset for metastasis detection in breast cancer, including 270 training WSIs and 130 test WSIs. The WSIs that contained metastasis were labeled positive, and the WSIs that did not contain metastasis were labeled negative. The dataset provides not only the labels of whether a WSI is positive but also pixel-level labels of the metastasis areas. Following the MIL paradigm, we used only slide-level labels in the training process and evaluated the performance of our method in both the slide-level classification task and the patch-level classification task. We divided each WSI into 512$\times$512 patches without overlapping under 5$\times$ magnification. Patches with an entropy less than 5 are discarded as background. A patch is labeled as positive only if it contains 25\% or more cancer areas; otherwise, it is labeled as negative. Eventually, a total of 186,604 patches were obtained, of which there were only 8117 positive patches (4.3\%).

\noindent\textbf{TCGA Lung Cancer dataset.} The TCGA Lung Cancer dataset includes a total of 1054 WSIs from The Cancer Genome Atlas (TCGA) Data Portal, which includes two sub-types of lung cancer, namely Lung Adenocarcinoma and Lung Squamous Cell Carcinoma. Only slide-level labels are available for this dataset, and patch-level labels are unknown. Following DSMIL \cite{9}, this dataset contains about 5.2 million patches at 20$\times$ magnification, with an average of about 5,000 patches per bag.

\subsection{Implementation Details}
For the MAE, we only modified its input size of the ViT model in its encoder and decoder (from 224 to 512), and other model details are the same as those in \cite{10}. For training the MAE, we set the mask ratio to 75\%, use the Adam optimizer with a learning rate of 1.5e-4, a weight decay of 0.05, a batch size of 128, and train 500 epochs using 4 Nvidia 3090 GPUs. The number of clusters in section 2.4 is set to 10 classes, and the proportion of extreme instances selected for refinement in section 2.5 are set to 10\%. For feature refinement, we use Adam optimizer with an initial learning rate of 0.01 and cosine descent by epoch. Both the Linear Projector Head and Classification Head are composed of one fully connected layer. Refinement convergence means that the decrease of the cross-entropy loss is below a small threshold in 10 consecutive epochs.

\subsection{Evaluation Metrics and Comparison Methods}
We report the AUC and FROC metrics for patch-level localization task and the AUC and Accuracy metrics for slide-level classification task. We compared our method with the latest methods in the field of MIL-based histopathological image analysis \cite{3,6,7,8,9}. Among them, Ab-MIL \cite{6}, Loss-based-MIL \cite{8} and DSMIL \cite{9} are embedding-based methods and RNN-MIL \cite{3} and Chikontwe-MIL \cite{7} are key-instance-based methods.

\vspace{-2em}
\newcommand{\tablestyle}[2]{\setlength{\tabcolsep}{#1}\renewcommand{\arraystretch}{#2}\centering\footnotesize}
\begin{table*}[htbp]
  \centering
  \caption{Results of patch-level and slide-level metrics on CAMELYON16 dataset and TCGA Lung Cancer dataset.}

  \subfloat[Metrics on CAMELYON16 dataset.]
  {
    \hspace{-1em}
    \centering
    \begin{minipage}[b]{0.5\linewidth}{
    \begin{center}
    \tablestyle{4pt}{1.05}
    \scalebox{0.49}{
        \begin{tabular}{ccc|cc}
            \toprule
            Method & Patch AUC & Patch FROC & Slide AUC & Slide Accuracy \\
            \midrule
            Ab-MIL (ICML’18) & 0.4739 & 0.1202 & 0.6612 & 0.6746 \\
            RNN-MIL (Nat. Med.’19) & 0.6055 & 0.3011 & 0.6913 & 0.6845 \\
            Loss-based-MIL (AAAI’20) & 0.6173 & 0.3012 & 0.7024 & 0.7011 \\
            Chikontwe-MIL (MICCAI’20) & 0.7880  & 0.3717 & 0.7024 & 0.7123 \\
            DSMIL (CVPR’21) & 0.8873 & 0.4560  & 0.7544 & 0.7359 \\
            \midrule
            \textbf{DGMIL (Ours)} & \textcolor[rgb]{ 1,  0,  0}{\textbf{0.9045}} & \textcolor[rgb]{ 1,  0,  0}{\textbf{0.4887}} & \textcolor[rgb]{ 1,  0,  0}{\textbf{0.8368}} & \textcolor[rgb]{ 1,  0,  0}{\textbf{0.8018}} \\
            \midrule
            \textcolor[rgb]{ .459,  .443,  .443}{Fully-supervised} & \textcolor[rgb]{ .459,  .443,  .443}{0.9644} & \textcolor[rgb]{ .459,  .443,  .443}{0.5663} & \textcolor[rgb]{ .459,  .443,  .443}{0.8621} & \textcolor[rgb]{ .459,  .443,  .443}{0.8828} \\
            \bottomrule
            \end{tabular}%
    }
    \end{center}
    }\end{minipage}
  }
  \subfloat[Metrics on TCGA Lung Cancer dataset.]
  {
    \centering
    \begin{minipage}[b]{0.5\linewidth}{
    \begin{center}
    \tablestyle{4pt}{1.05}
    \scalebox{0.49}{
        \begin{tabular}{ccc}
            \toprule
            Method & Slide AUC & Slide Accuracy \\
            \midrule
            Mean-pooling & 0.9369 & 0.8857 \\
            Max-pooling & 0.9014 & 0.8088 \\
            Ab-MIL (ICML’18) & 0.9488 & 0.9000  \\
            RNN-MIL (Nat. Med.’19) & 0.9107  & 0.8619 \\
            DSMIL (CVPR’21) & 0.9633 & 0.9190  \\
            \midrule
            \textbf{DGMIL (Ours)} & \textcolor[rgb]{ 1,  0,  0}{\textbf{0.9702}} & \textcolor[rgb]{ 1,  0,  0}{\textbf{0.9200 }} \\
            \bottomrule
            \end{tabular}%
    }
    \end{center}
    }\end{minipage}
  }
  \label{table1}%
\end{table*}%



\vspace{-2em}
\subsection{Results}
Table \ref{table1} (a) shows the performance of our method and the competitors for both slide-level and patch-level classification tasks on CAMELYON16 dataset, and our method achieves the best performance on both tasks. Ab-MIL \cite{6} used bag-level loss and used attention mechanism for patch classification, which can hardly classify the patches. Loss-based-MIL \cite{8} utilized both a slide-level loss and a patch-level loss, and it achieved higher performances than Ab-MIL on both tasks. Both RNN-MIL \cite{3} and Chikontwe-MIL \cite{7} used the output probability of the instance classifier in the current iteration to select the key instances. However, the pseudo labels for some selected key instances may be wrong, and the true positive instances may not be fully selected, which lead to their limited performance. DSMIL \cite{9} uses the contrastive self-supervised framework Simclr \cite{14} for instance-level feature extraction and uses Masked Non-Local Operation-based dual-stream aggregator for both instance-level and bag-level classification, and it achieved the current SOTA performance on both tasks. However, its loss function is also defined at the bag-level and without extra guidance of data distribution, and the model is not motivated to make accurate predictions of all instances.
In comparison, our proposed method DGMIL models the data distribution properly through iterative refinement of the feature space and achieve the highest performance on both tasks. Especially, the slide-level AUC of DGMIL is significant higher than that of DSMIL.

Table \ref{table1} (b) shows the performance of our method and the competitors for the slide-level classification task on TCGA Lung Cancer dataset, and our method also achieves the best performance.

\vspace{-2em}
\begin{table*}[htbp]
    \centering
    \caption{Results of ablation tests on CAMELYON16 dataset.}
  
    \subfloat[Ablation tests of main strategies.]
    {
      \hspace{-1em}
      \centering
      \begin{minipage}[b]{0.5\linewidth}{
      \begin{center}
      \tablestyle{4pt}{1.05}
      \scalebox{0.55}{
          \begin{tabular}{cccccc}
              \toprule
              \multirow{2}[2]{*}{Method} & Distribution  & One-shot  & Iterative & Patch & Slide  \\
                    & Modeling & Refinement &  Refinement & AUC   & AUC \\
              \midrule
              Baseline1 &       &       &       & 0.7672 & 0.7637 \\
              Baseline2 & \textbf{\checkmark} &       &       & 0.8371 & 0.7355 \\
              one-shot DGMIL & \textbf{\checkmark} & \textbf{\checkmark} &       & 0.8899 & 0.8267 \\
              DGMIL & \textbf{\checkmark} &       & \textbf{\checkmark} & \textcolor[rgb]{ 1,  0,  0}{\textbf{0.9045}} & \textcolor[rgb]{ 1,  0,  0}{\textbf{0.8368}} \\
              \bottomrule
              \end{tabular}%
      }
      \end{center}
      }\end{minipage}
    }
    \subfloat[Further tests on feature space refinement.]
    {
      \centering
      \begin{minipage}[b]{0.5\linewidth}{
      \begin{center}
      \tablestyle{4pt}{1.05}
      \scalebox{0.49}{
          \begin{tabular}{ccc}
              \toprule
              Ratio & Patch AUC & Slide AUC \\
              \midrule
              1\%   & 0.8443 & 0.7649 \\
              5\%   & 0.8903 & 0.8293 \\
              10\%  & \textcolor[rgb]{ 1,  0,  0}{\textbf{0.9045}} & \textcolor[rgb]{ 1,  0,  0}{\textbf{0.8368}} \\
              20\%  & 0.8911 & 0.8196 \\
              30\%  & 0.8897 & 0.8280 \\
              Existing SOTA & 0.8873 & 0.7544 \\
              \bottomrule
              \end{tabular}%
      }
      \end{center}
      }\end{minipage}
    }
    \label{table2}%
  \end{table*}%

\vspace{-2em}
\section{Ablation Study}
Table \ref{table2} (a) shows the results of our ablation tests on the CAMELYON16 dataset. \textbf{Baseline1} denotes that instead of using Cluster-conditioned Feature Distribution Modeling, an instance-level classifier is directly trained to predict all instances using a key-instance-based training approach similar to \cite{3,8}, where the key instances are selected according to the top 10\% instances with the largest output probability and the top 10\% instances with the smallest output probability. \textbf{Baseline2} denotes that only the initial feature space obtained by MAE is used to model feature distribution and calculate positive scores without feature space refinement. \textbf{One-shot DGMIL} indicates that the feature space refinement is done only once.

It can be seen that \textbf{Baseline1} does not achieve good performance because the key instances selected by the network are likely to be inaccurate. In addition, the direct use of the initial feature space based on MAE does not model the distribution of positive instances and negative instances very well, because the training of MAE is completely self-supervised and lacks the guidance of bag-level supervision information. Therefore, the initial feature space is not fully suitable for distribution modeling and the performance of \textbf{Baseline2} is not very high. On the other hand, one-shot refinement of the feature space (\textbf{one-shot DGMIL}) can improve the performance and multiple iterative refinement help DGMIL achieve the best performance.

Table \ref{table2} (b) shows the effect of the proportion of selected extreme instances on the final results during iterative feature space refinement. When too few extreme instances are selected, the feature space cannot be adjusted adequately. On the contrary, when too many extreme instances are selected, due to the low proportion of positive patches in the dataset itself, the number of false pseudo labels increases, resulting in distortion of the feature space. It can be seen that 10\% is the optimal choice. At the same time, these results indicate that the proposed method is robust to the choice of the proportion of extreme instances. Using any proportion from 1\% to 30\%, the slide-level AUCs all exceed the current SOTA, and using any proportion from 5\% to 30\%, the patch-level AUCs all exceed the current SOTA.

\begin{figure*}[htbp]
    \centering
    \includegraphics[scale=0.23]{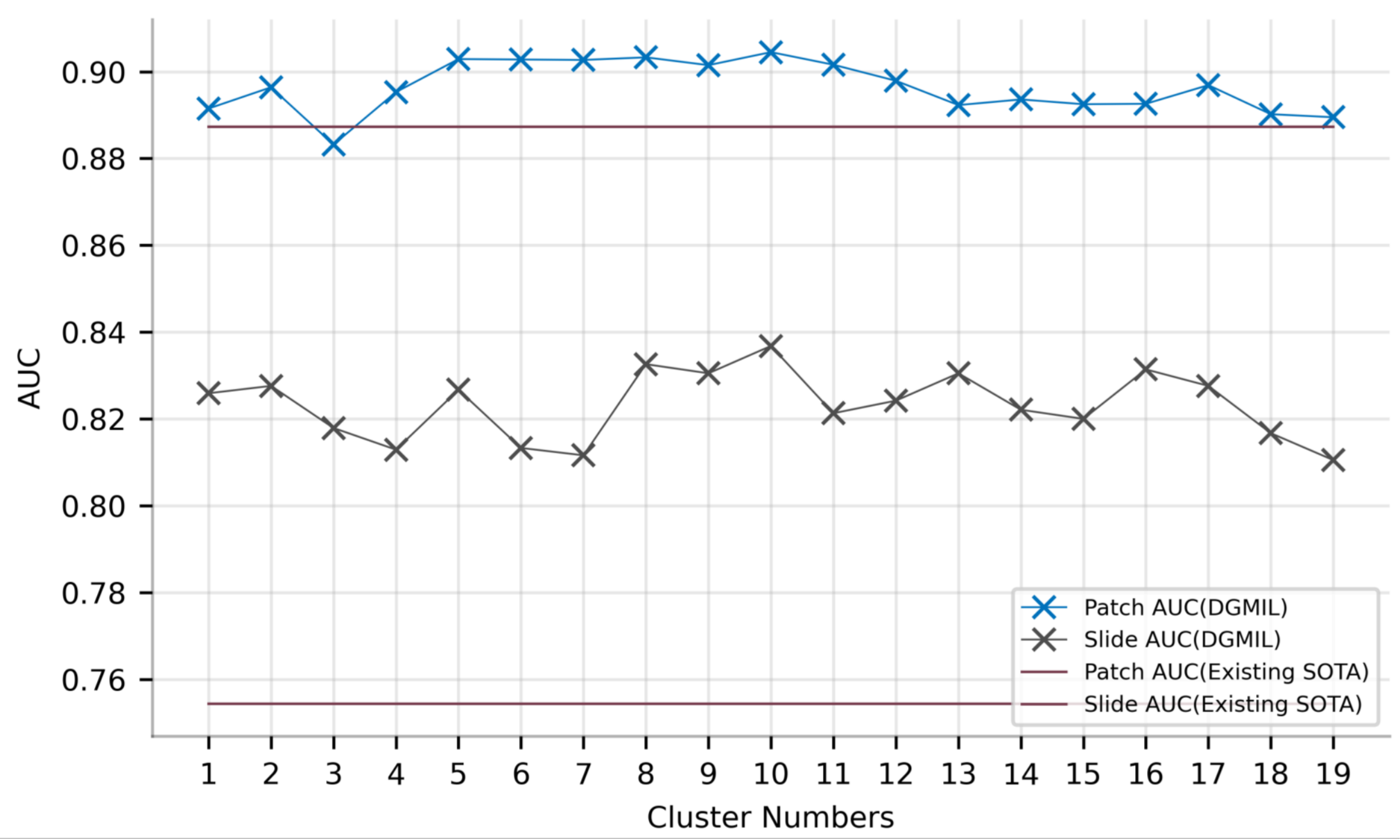}
    \caption{Results of ablation tests on the number of clusters in Feature Distribution Modeling.}
    \label{figure2}
\end{figure*}

Fig. \ref{figure2} shows the effect of the number of clusters on the performance when performing feature distribution modeling, where a cluster number of one indicates that no clustering is performed. As can be seen, our method is not sensitive to the number of clusters, but performing clustering does work better than not performing clustering. The reason is that many different phenotypes exist in negative instances, such as normal cells, blood vessels, fat, and others in normal tissues, so the distribution of negative instances is fairly sparse. Clustering can divide different phenotypes into different clusters and help make the instance-to-cluster distance better reflect whether an instance is negative or not. It can be seen that clustering into 10 classes is the most effective choice.

\section{Conclusions}
In this paper, we propose DGMIL: a feature distribution guided deep MIL framework for WSI classification and positive patch localization. We propose a cluster-conditioned feature distribution modeling method and a pseudo label-based iterative feature space refinement strategy. Instead of designing complex discriminative classifiers, we construct a proper feature space in which the positive and negative instances can be easily separated. New SOTA performance is achieved on the CAMELYON16 public breast cancer metastasis detection dataset and the TCGA Lung Cancer dataset.

\section*{Acknowledgments}

This work was supported by National Natural Science Foundation of China under Grant 82072021. The TCGA Lung Cancer dataset is from the
TCGA Research Network: https://www.cancer.gov/tcga.

\bibliographystyle{splncs04}
\bibliography{miccai2022}
\end{document}